\def\year{2021}\relax
\documentclass[letterpaper]{article} 
\usepackage{aaai21}  
\usepackage{times}  
\usepackage{helvet} 
\usepackage{courier}  
\usepackage[hyphens]{url}  
\usepackage{graphicx} 
\usepackage{natbib}  
\usepackage{caption} 
\usepackage{amsmath}
\usepackage{amssymb}
\usepackage{bm}
\usepackage[italicdiff]{physics}
\DeclareMathOperator*{\expectation}{\mathbb{E}}
\usepackage{booktabs}
\usepackage[labelformat=simple]{subcaption}

\newtheorem{theorem}{Theorem}
\title{Accelerating Continuous Normalizing Flow with Trajectory Polynomial Regularization}
\author {
        Han-Hsien Huang,\textsuperscript{\rm 1,2}
        Mi-Yen Yeh \textsuperscript{\rm 1} \\
}
\affiliations {
    \textsuperscript{\rm 1} Institute of Information Science, Academia Sinica \\
    \textsuperscript{\rm 2} Department of Computer Science and Engineering, Texas A\&M University \\
    hanhsien.huang@tamu.edu, miyen@iis.sinica.edu.tw
}
\begin{document}

\maketitle

\begin{abstract}
In this paper, we propose an approach to effectively accelerating the computation of continuous normalizing flow (CNF), which has been proven to be a powerful tool for the tasks such as variational inference and density estimation. The training time cost of CNF can be extremely high because the required number of function evaluations (NFE)  for solving corresponding ordinary differential equations (ODE) is very large. We think that the high NFE results from large truncation errors of solving ODEs. To address the problem, we propose to add a regularization. The regularization penalizes the difference between the trajectory of the ODE and its fitted polynomial regression. The trajectory of ODE will approximate a polynomial function, and thus the truncation error will be smaller. Furthermore, we provide two proofs and claim that the additional regularization does not harm training quality. Experimental results show that our proposed method can result in 42.3\%  to  71.3\% reduction of NFE on the task of density estimation, and 19.3\%  to  32.1\% reduction of NFE on  variational auto-encoder, while the testing losses are not affected.
\end{abstract}

\section{Introduction}\label{introduction}
Normalizing flows \citep{rezende2015variational} are a kind of invertible neural networks that have an efficient calculation of Jacobian determinant.
They can be used as generative models, density estimation or posterior distribution of variational auto-encoders (VAE).
However, their two requirements, invertibility and easy Jacobian determinant calculation, put a great restriction on the design of corresponding neural network architecture. 

Recently, continuous normalizing flow (CNF) \citep{chen2018neural,grathwohl2018ffjord} is proposed that can avoid such design restriction. CNF describes the transformation of hidden states with ordinary differential equations (ODE), instead of layer-wise function mappings. In this way, the invertibility becomes trivial and the determinant of Jacobian becomes simply a trace. With the freedom of network architecture design, CNF has been shown to outperform normalizing flow with discrete layers in terms of lower testing loss.

However, running a CNF model, which is equal to solving a series of ODEs, may take a lot of time. 
A single training iteration can need up to hundreds of function evaluations, which is equivalent to running a neural network with hundreds of layers.
Moreover, the number of function evaluations required per training iteration can gradually grow up throughout the training process.
The reason is that CNF uses ODE solvers with adaptive steps.
The step sizes of solving ODEs are determined based on the truncation errors. Such error usually grows as the training goes and thus the number of steps becomes larger.

To make training of CNF faster, we propose \emph{Trajectory Polynomial Regularization} (TPR).
TPR is an additional loss function which regularizes the trajectories of the ODE solutions. It penalizes the difference between the trajectory of solution and its fitted polynomial regression. Therefore, TPR enforces the solutions to approximate polynomial curves, which can reduce the truncation error of solving the ODEs of CNF. 
Despite adding the regularization, we argue that the method does not harm the testing loss much. 
We prove that the optimal solutions of the modified loss are still optimal solutions for the original task. The detail of the proofs and argument are in Section~\ref{sec:method}.
In experiments, we find that for both density estimation and variational inference tasks, 
our model can save as much as 71\% of \emph{number of function evaluations} (NFE), which is the number of evaluating the ODEs.
Furthermore, our method barely affect the testing errors.
Our code is published in our Github repository\footnote{\url{https://github.com/hanhsienhuang/CNF-TPR}}.

The remainder of the paper is organized as follows.
We first introduce the background and related work of CNF in Section \ref{sec:prelimiaries}.
In Section \ref{sec:method}, we describe the intuition and mathematics of TPR and provide two proofs to argue the power of its quality.
In Section \ref{sec:experiments}, we conduct experiments on two tasks, density estimation and VAE. We visualize the effect of our model with simple 2D toy data. And then we evaluate the efficiency and quality of our model on real data.

\section{Preliminaries}\label{sec:prelimiaries}
\subsection{Background}\label{sec:background}
Normalizing flows \citep{rezende2015variational} are models that can be used to represent a wide range of distributions. They consist of an invertible function.
Let $f:\mathbb{R}^d\to\mathbb{R}^d$ be an invertible function, i.e., $\vb{z}=f(\vb{x})$ and $\vb{x}=f^{-1}(\vb{z})$.
Due to the invertibility, the log probabilities of the input $\vb{x}$ and the output $\vb{z}=f(\vb{x})$ have the following relation.
\begin{equation}\label{eq:nf}
    \log p(\vb{z})=\log p(\vb{x}) - \log \abs{\det \pdv{f(\vb{x})}{\vb{x}}}.
\end{equation}
Once $p(\vb{x})$ is set to be a known distribution, the distribution $p(\vb{z})$ is represented by $p(\vb{x})$ and the function $f$. The function $f$ can be parametrized by deep neural networks. However, since $f$ should be invertible and the Jacobian $\det\pdv{f}{\vb{x}}$ in equation \eqref{eq:nf} should be easily calculated, the possible architecture of $f$ is highly restricted. As a result, the capacity of representing $p(\vb{x})$ is also limited.

Different from normalizing flows, continuous normalizing flow (CNF) \citep{chen2018neural} makes the transformation from $\vb{x}$ to $\vb{z}$ a continuous evolution.
That is, instead of a direct function mapping from $\vb{x}$ to $\vb{z}$, i.e.\ $\vb{z}=f(\vb{x})$, the transformation can be represented by a continuous function $\vb{y}(t)$, with $\vb{x}=\vb{y}(t_0)$ and $\vb{z}=\vb{y}(t_1)$. 
The evolution of $\vb{y}(t)$ in CNF is defined by an ordinary differential equation (ODE),
\begin{equation}
\dv{\vb{y}}{t} = \vb{v}(\vb{y}, t).
\label{eq:dynamics_y}
\end{equation}
With this continuous transformation, the invertibility of $\vb{x}$ and $\vb{z}$ is inherently satisfied. Therefore, there isn't any other restriction on the form of function $\vb{v}(\vb{y}, t)$. 

The dynamics of the log likelihood $\log p(\vb{y}(t))$ is derived in the instantaneous change of variables theorem by \citet{chen2018neural}, which is
\begin{equation}
\dv{t} \log p = -\Tr(\pdv{\vb{v}}{\vb{y}}).
\label{eq:dynamics_logp}
\end{equation}
The computation complexity for obtaining the exact value of $\Tr(\pdv{\vb{v}}{\vb{y}})$ is $\order{D^2}$, where $D$ is the dimension of data $\vb{x}$. 
It can be reduced to $\order{D}$ \cite{grathwohl2018ffjord} with the Hutchinson's trace estimator \citep{hutchinson1989stochastic}.
\begin{equation}\label{eq:hutchinson}
    \Tr(\pdv{\vb{v}}{\vb{y}})=\expectation_{p(\bm{\epsilon})}\left[\bm{\epsilon}^T\pdv{\vb{v}}{\vb{y}}\bm{\epsilon}\right],
\end{equation}
where $\expectation[\bm{\epsilon}]=0$ and $\mathrm{Cov}(\bm{\epsilon})=I$.
With this method, computing the estimation of $\Tr(\pdv{\vb{v}}{\vb{y}})$ costs exactly the same as computing $\vb{v}$.

Given the initial values of $\vb{y}(t_0)$ and $\log p(t_0)$, CNF outputs the final values $\vb{y}(t_1)$ and $\log p(t_1)$, which can be written as
\begin{align}\label{eq:int_y}
        \vb{y}(t_1) &= \vb{y}(t_0) + \int^{t_1}_{t_0} \vb{v} \dd{t}, \text{and}
        \\
        \label{eq:int_logp}
        \log p(\vb{y}(t_1)) &= \log p(\vb{y}(t_0)) - \int^{t_1}_{t_0}\Tr(\pdv{\vb{v}}{\vb{y}}) \dd{t}.
\end{align}
These equations are solved by numerical ODE solvers.
Usually, adaptive stepsize solvers, such as Dopri5 \citep{dormand1980family}, are used because they guarantee a desired precision of output. 
Adaptive solvers dynamically determine the step sizes of solving ODE by estimating the magnitude of truncation errors.
They shrink the step sizes when the truncation errors exceed a tolerable value and increase the step sizes when the errors are too small.

Unlike the restriction of $f$'s architecture in normalizing flows, the $\vb{v}(\vb{y}, t)$ in CNF can be represented by arbitrary neural networks. 
As a result, CNF can have greater capacity to represent the distributions $p(\vb{z})$ than other normalizing flow models. 

\subsection{Related work}

The design of normalizing flows is  restricted by the requirements of invertibility and efficient Jacobian determinant computation.
\citet{rezende2015variational} proposed planar flow and radial flow for variational inference. The layers in the proposed model have to be in a simple form.
Real NVP \citep{dinh2016density} is a structure of network such that its Jacobian matrix are triangular, so the determinant are simply the product of diagonal values.
\citet{kingma2018glow} proposed Glow, which has $1\times 1$ convolution layers that pass partitioned dimensions into invertible affine transformations. 
Auto-regressive flow models \citep{kingma2016improved, papamakarios2017masked, oliva2018transformation, huang2018neural} also have triangular Jacobian matrix in each layer. Although they have high expressiveness, they require $D$ sequential computations in each layer, which can be costly when $D$ is large.
\citet{huang2020augmented} proposed Augmented Normalizing Flow (ANF), which augments the data by adding more dimensions in transformation.  Although reaching high performance, it requires large number of samples to compute accurate importance sampling.

Continuous normalizing flow (CNF) \citep{chen2018neural} is a continuous version of normalizing flows, which replaces the layer-wise transformations of normalizing flows with ODEs. 
The model does not restrict the structure of its transformation function, but computing Jacobian determinant costs computing gradient $D$ times.
\citet{grathwohl2018ffjord} improved the efficiency of computing Jacobian determinant by using Hutchinson's trace estimator. 
\citet{chen2019neural} proposed a new architecture of neural network which can efficiently compute trace of Jacobian.

As we mentioned in Section \ref{sec:background}, training CNF takes a lot of time because solving ODEs can be very slow. To enhance the training stability and reduce the training time, some researchers have proposed to adopt regularization methods. \citet{grathwohl2018ffjord} proposed to use weight decay and spectral normalization \citep{miyato2018spectral}.
\citet{salman2018deep} proposed geodesic regularization and inverse consistency regularization. \citet{finlay2020train} and \citet{onken2020ot} proposed to regularize the dynamics of CNF with the notion of optimal transport.
\citet{kelly2020learning} proposed to regularize the higher-order derivatives of the velocity field and applied Taylor-mode automatic differentiation to compute the derivatives efficiently. 
Our work is different from these works in two ways. First, we provide proofs for the existence of infinite optimal solutions with our regularization. Second, our proposed TPR only requires a few samples point in time. On the contrary, those methods need to solve more than one additional ODEs, which increases computational consumption.

\section{Method}\label{sec:method}
\begin{figure*}
    \centering
    \begin{subfigure}[t]{0.38\textwidth}
        \includegraphics[width=\linewidth]{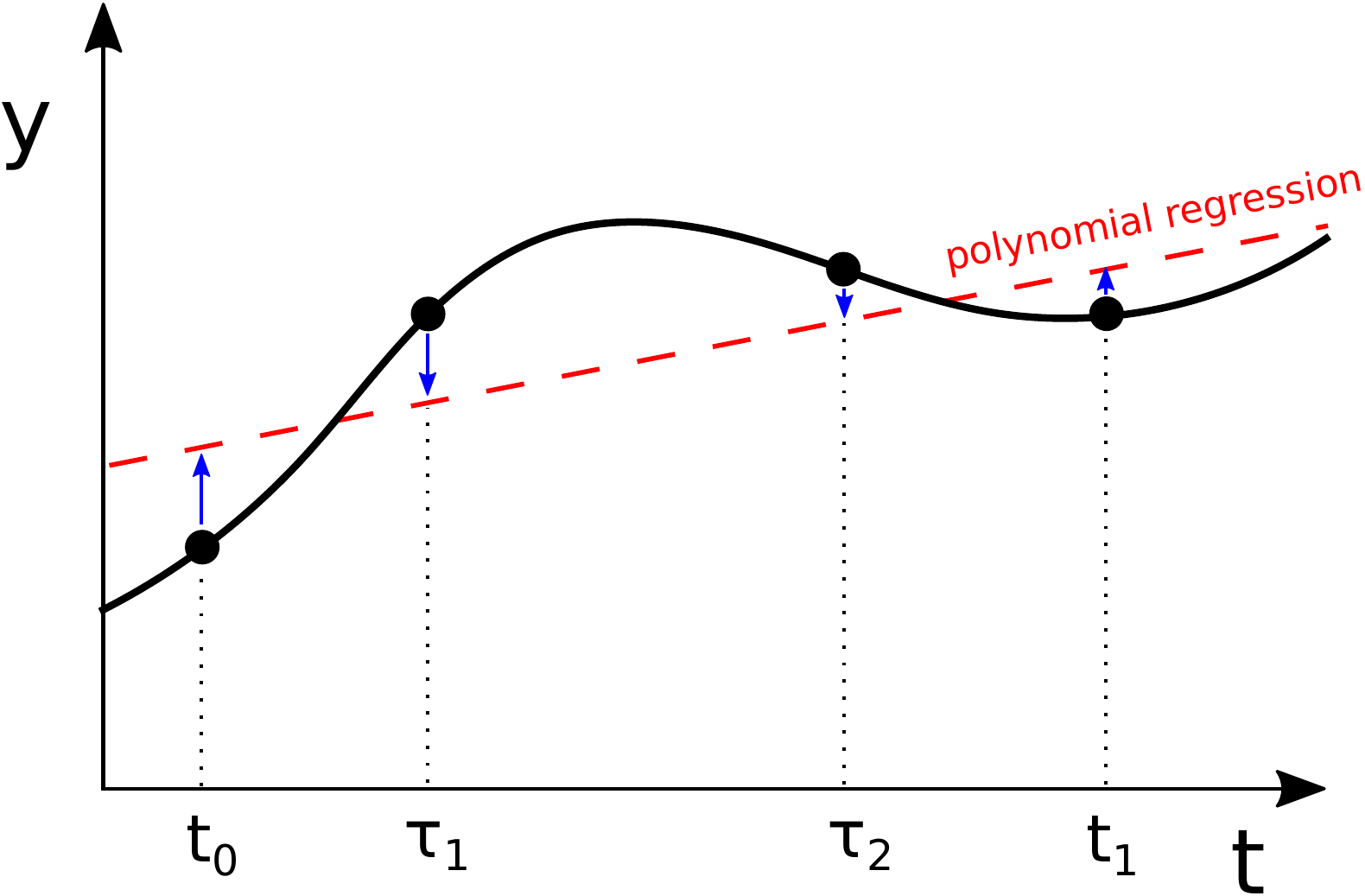}
        \caption{Illustration of the mechanism of TPR}
        \label{subfig:TPR}
    \end{subfigure}
    \begin{subfigure}[t]{0.3\textwidth}
        \includegraphics[width=\linewidth]{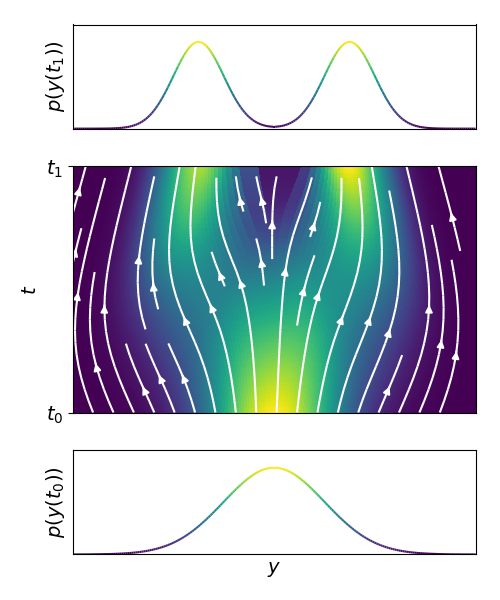}
        \caption{Without TPR.}
        \label{subfig:withoutTPR}
    \end{subfigure}
    \begin{subfigure}[t]{0.3\textwidth}
        \includegraphics[width=\linewidth]{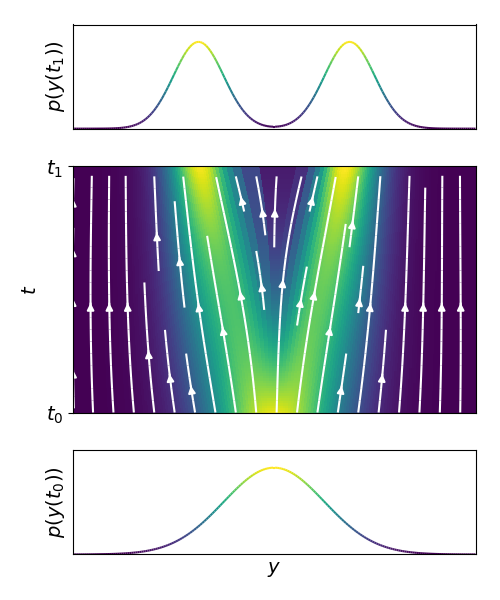}
        \caption{With TPR.}
        \label{subfig:withTPR}
    \end{subfigure}
    \caption{Illustration of our Trajectory Polynomial Regularization (TPR). \subref{subfig:TPR}: The illustrated mechanism of TPR. The solid curve represents trajectory of $\vb{y}(t)$. The dashed line is a fitted polynomial regression on the four randomly sampled points. Polynomial regression of degree 1 is demonstrated here. The TPR loss $L_p$ is designed to pull the points of $\vb{y}$ to the fitted polynomial. 
     \subref{subfig:withoutTPR} and \subref{subfig:withTPR}: Two transformations of 1D distributions by CNF either with or without our TPR. The input is a Gaussian distribution and the output is a mixture of two Gaussian distributions. The white streamlines represent the trajectories of $\vb{y}(t)$ and the color represents the density $p(\vb{y}(t))$. }
    \label{fig:tpr}
\end{figure*}

The inefficiency of CNF comes from the large number of steps of solving ODEs. With adaptive ODE solvers, the number of steps is dynamic. The solvers decrease the step sizes when the truncation errors of solving OEDs are too large, and vice versa.
Therefore, to reduce the number of steps, we propose a method to reduce the truncation errors when solving ODEs. 
To do so, we focus on reducing the truncation error of solving $\vb{y}(t)$ but not $\log p$.
Although we actually have two ODEs to solve (equations of \eqref{eq:dynamics_y} and \eqref{eq:dynamics_logp}), both their right-hand-sides depend only on $\vb{y}$.

Theoretically, for a $k$-th order ODE solver to solve the equation $\dv{\vb{y}}{t}=\vb{v}(\vb{y},t)$, the local truncation error of $\vb{y}$ is $\order{\dv[k+1]{\vb{y}}{t}\Delta t^{k+1}}$, where $\Delta t$ is the size of step. To increase the step size $\Delta t$ while maintaining the magnitude of truncation error, $\norm{\dv[k+1]{\vb{y}}{t}}$ should be as small as possible. Obviously, the optimal situation is that $\vb{y}(t)$ is a polynomial function of degree $d\le k$, because $\dv[k+1]{\vb{y}}{t}=0$.

To force the trajectory $\vb{y}(t)$ to approximate a polynomial function, we add a loss term $L_p$ called \emph{Trajectory Polynomial Regularization} (TPR) into the total loss. 
\begin{equation}
    L = L_0 + \alpha L_p,
    \label{eq:total_loss}
\end{equation}
where $L_0$ is the original loss function of the corresponding task and $\alpha$ is a constant coefficient. $L_p$ is derived as follows. First, a number of time steps $\{\tau_0, \dots, \tau_{n-1}\}$ are randomly sampled. Next, a polynomial function $\vb{f}(t)$ is fitted to the points $\{\vb{y}(\tau_0),\dots,\vb{y}(\tau_{n-1})\}$. And finally, $L_p$ is calculated as the mean squared error (MSE) between $\vb{f}(\tau_i)$ and $\vb{y}(\tau_i)$. When minimizing $L_p$, $\vb{y}(t)$ will approach the polynomial function, $\vb{f}(t)$.

Figure~\ref{fig:tpr} illustrates the mechanism and effect of our method. Figure~\ref{subfig:TPR} demonstrates how TPR regularizes $\vb{y}(t)$ to approximate a polynomial function. 
The loss $L_p$ is defined as the squared difference between the two curves. So it act as a force to pull the four points from the trajectory to the fitted polynomial regression.
Figure~\ref{subfig:withoutTPR} shows a possible result without TPR. The trajectories of the solutions $\vb{y}(t)$ can be very winding, so solving $\vb{y}(t)$ can be more difficult. 
Because the truncation errors are higher when solving $\vb{y}(t)$, ODE solvers then have to shrink step sizes in order to guarantee the precision of the solution, which leads to more computation in terms of higher number of function evaluations. 
On the other hand, Figure~\ref{subfig:withTPR} shows the result using TPR with degree 1. The trajectories of $\vb{y}(t)$ are more like a straight line. As a result, $\vb{y}(t)$ is easier to solve, and thus the required number of function evaluations decreases.

To derive $L_p$, the math expression of $L_p$ can be written as
\begin{equation}\label{eq:polynomial_mse}
    L_p = \frac{1}{n} \sum_{i=0}^{n-1} \norm{ \vb{y}(\tau_i)-\vb{f}(\tau_i)}^2 
    = \frac{1}{n}\norm{Y-TC}^2,
\end{equation}
where $\norm{\cdot}$ is Frobenius norm. 
$\vb{f}(t) = \vb{c}_0 P_0(t) + \cdots + \vb{c}_d P_d(t)$ is the polynomial function we want to fit. $d$ is the degree of the polynomial function, $\{\vb{c}_0, \dots , \vb{c}_d\}$ is the coefficients to be fitted and $\{P_0(t),\dots,P_d(t)\}$ is a polynomial basis. The matrices $Y$, $T$ and $C$ are written as follows,
\begin{equation}
    Y = \left(
    \begin{matrix}
    \vb{y}(\tau_0)& \cdots & \vb{y}(\tau_{n-1})
    \end{matrix}
    \right)^\top,
\end{equation}
\begin{equation}
    T = 
    \left(
    \begin{matrix}
        P_0(\tau_0) & \cdots & P_d(\tau_0) \\
        \vdots & \ddots & \vdots \\
        P_0(\tau_{n-1}) & \cdots & P_d(\tau_{n-1})
    \end{matrix}
    \right),
\end{equation}
\begin{equation}
    C = \left(
    \begin{matrix}
    \vb{c}_0 & \cdots & \vb{c}_d 
    \end{matrix}
    \right)^\top.
\end{equation}
Solving $C$ to minimize $\norm{Y-TC}^2$ yields to the equation, $C = (T^\top T)^{-1}T^\top Y$. After substituting it back to equation \eqref{eq:polynomial_mse}, we can get
\begin{equation}
    L_p = \frac{1}{n}\norm{\left(I-T(T^\top T)^{-1} T^\top\right) Y}^2.
\end{equation}
To let the matrix computation more numerically stable, we use singular value decomposition (SVD) on $T$, which gives us $T=U\Sigma V^\top$. And thus, the final form of calculating the TPR loss is written as follows,
\begin{equation}\label{eq:polynomial_final}
    L_p = \frac{1}{n} \norm{ \left(I-U_{1:d+1}(U_{1:d+1})^\top\right)Y }^2, 
\end{equation}
where $U_{1:d+1}$ denotes the leftmost $d+1$ columns of matrix $U$.

Although we introduce a new loss term into the total loss, we prove that there are always optimal solutions minimizing both $L$ and $L_0$.
In other words, when minimizing the total loss $L$ to find an optimal solution, the solution is also an optimal solution of $L_0$.
In the following two theorems, we prove that, even with the hard constraint $\dv[2]{\vb{y}}{t} = \dv{\vb{v}}{t}=0$ (so $L_p=0$), there still exist infinitely many functions $\vb{v}(\vb{y}, t)$ that can transform any distribution to any other distribution. 
Because $L_0$ is a function of $\log p(\vb{y}(t_1))$, from the theorems, there are infinitely many $\vb{v}(\vb{y}, t)$ that can simultaneously minimize $L_0$ and $L_p$. Therefore, these $\vb{v}$ can minimize $L$ and $L_0$ simultaneously. 
We present the two theorems as follows.

\begin{theorem}
  Assume that $\vb{y}(t)\in \mathbb{R}^D$ and $\log p(\vb{y}(t))$ are governed by the differential equations \eqref{eq:dynamics_y} and \eqref{eq:dynamics_logp}, respectively.
  Given any distributions $p_0(\vb{x})$ and $p_1(\vb{x})$, where $p_0(\vb{x})$ and $p_1(\vb{x}) > 0$ for all $\vb{x}\in\mathbb{R}^D$,  
  there exists a vector field $\vb{v}(\vb{y}, t)$ with the constraint $\dv{\vb{v}}{t}=0$ everywhere, such that if the initial value of  $\log p(\vb{y}(t))$ is $\log p_0(\vb{y}(t_0))$, then its final value is $\log p_1(\vb{y}(t_1))$.
  \label{thm:existence}
\end{theorem}
\begin{theorem}
  For $D>1$, there are infinitely many such vector fields $\vb{v}(\vb{y},t)$.
  \label{thm:infinity}
\end{theorem}

The next question is whether we can approach the optimal solutions with a function approximator, which is usually neural networks. We can't prove or guarantee anything about that, but we provide two arguments. 
First, the number of optimal solutions is infinite. Compared to finite number, the neural networks are more likely to approach one of the infinitely many optimal solutions.
Second, in experiments, we will show that our method doesn't affect testing loss much.  The model with TPR has approximately the same testing loss as the model without TPR.

The details of the proofs are in Appendix \ref{appendix:proofs} (see supplementary files).
The basic idea to prove Theorem~\ref{thm:existence} is to actually construct a vector field $\vb{v}(\vb{y},t)$ and 
the solution $\vb{y}(t)$ that satisfy all constraints.
The two main constraints in the theorem should be carefully considered.
First, the solution should satisfy the initial and final values.
Second, $\vb{v}(\vb{y}, t)$ must be a well-defined function at any points $\vb{y}$ and $t$.
For Theorem~\ref{thm:infinity}, the idea is to find an infinitesimal variation of $\vb{v}$ such that all the conditions are still satisfied. 
It should be careful that the value of the variation should be bounded in all space.

\section{Experiments}\label{sec:experiments}

We evaluate our proposed trajectory polynomial regularization (TPR) on two tasks, density estimation and variational autoencoder (VAE) \citep{kingma2013auto}. We compare our results with FFJORD \citep{grathwohl2018ffjord}, the current state of the art of CNF, and the corresponding state-of-the-art normalizing flow models. 

Two metrics are evaluated, testing loss and number of function evaluations (NFE). We want to see whether our model leads to lower computational cost compared to FFJORD in training, while keeping comparable training quality. To compare the computational cost, we count the average NFE per training iteration. NFE is defined as the number of evaluating the right-hand-side of the ODEs \eqref{eq:dynamics_y} and \eqref{eq:dynamics_logp} when solving them. The lower NFE, the less computational cost.
 
In all experiments, We use exactly the same architectures and hyper-parameters of neural network as in FFJORD. 
Also the same as FFJORD, the package of ODE solvers  Torchdiffeq\footnote{\url{https://github.com/rtqichen/torchdiffeq}} \citep{chen2018neural} is used. 
The only thing we do is adding our TPR loss to training loss.
The hyper-parameters of TPR are described below. We choose the number of sampled point as $n=4$, with $\tau_0 = t_0$ and $\tau_{n-1}=t_1$ as the start and end of time steps. The other $\tau$s are randomly sampled from a uniform distribution. The degree of the polynomial regression is set to be $d=1$. We adopted the same ODE solver as FFJORD~\cite{grathwohl2018ffjord}, which is Dopri5~\cite{dormand1980family}. Dopri5 is a 4th order ODE solver, so we can use any degree with $d\le 4$. In this work, we only choose $d=1$ in experiments to prove our concept. 
Despite the seemingly small degree, we have showed in Section \ref{sec:method} that it is powerful enough because the optimal solutions of $L_0$ still exist.
The coefficient of the polynomial regularization is $\alpha=5$. The tolerance coefficient of Dopri5 is set to be $atol=rtol=10^{-4}$ in training, and $atol=rtol=10^{-5}$ in testing to ensure precision of solving ODEs. These hyper-parameters above are shared in all experiments.

\subsection{Density estimation}
Given a data distribution $p(\vb{x})$, the task of density estimation aims at approximating this distribution.
Let $q_\theta(\vb{x})$ be the approximated distribution.
The loss of this task is the negative log-likelihood shown below.
\begin{equation}
    L_0 = -\expectation_{p(\vb{x})} \log q_\theta(\vb{x}).
\end{equation}
We experiment density estimation on three 2D toy data and real data sets, which are
five tabular data sets from \citep{papamakarios2017masked} and two image data sets, MNIST and CIFAR10.

\subsubsection{Results on 2D toy data}
\newcommand\picsize{2.6cm}
\begin{figure*}
    \centering
    \begin{subfigure}{0.18\linewidth}
        \centering
        \includegraphics[width=\picsize]{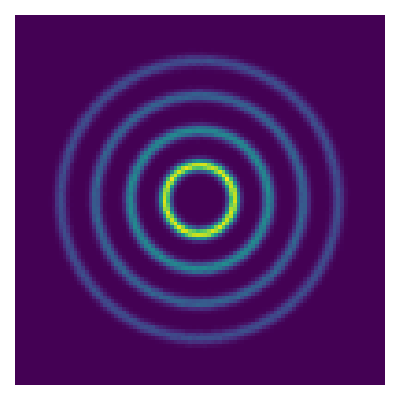}
        \includegraphics[width=\picsize]{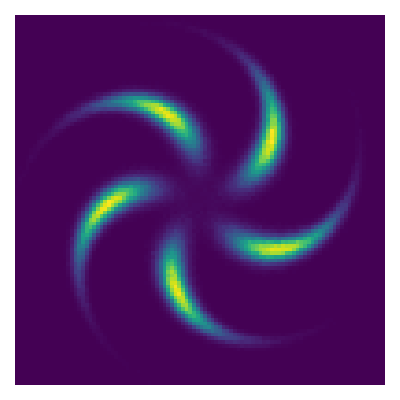}
        \includegraphics[width=\picsize]{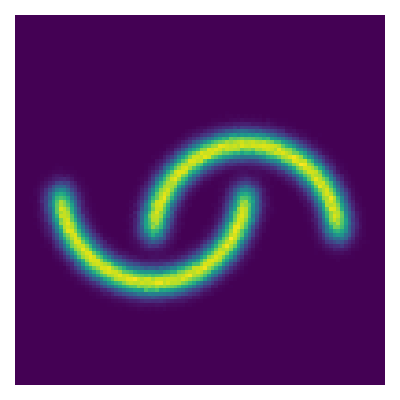}
        \caption{Data}
        \label{subfig:2d-data}
    \end{subfigure}
    \begin{subfigure}{0.33\linewidth}
        \centering
        \includegraphics[width=\picsize]{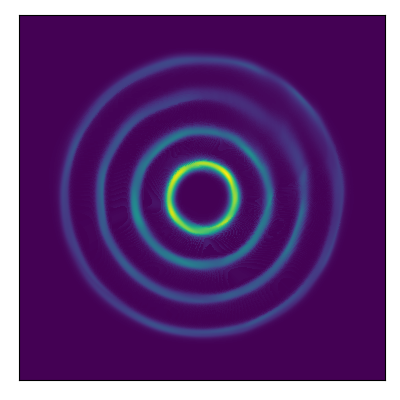}
        \includegraphics[width=\picsize]{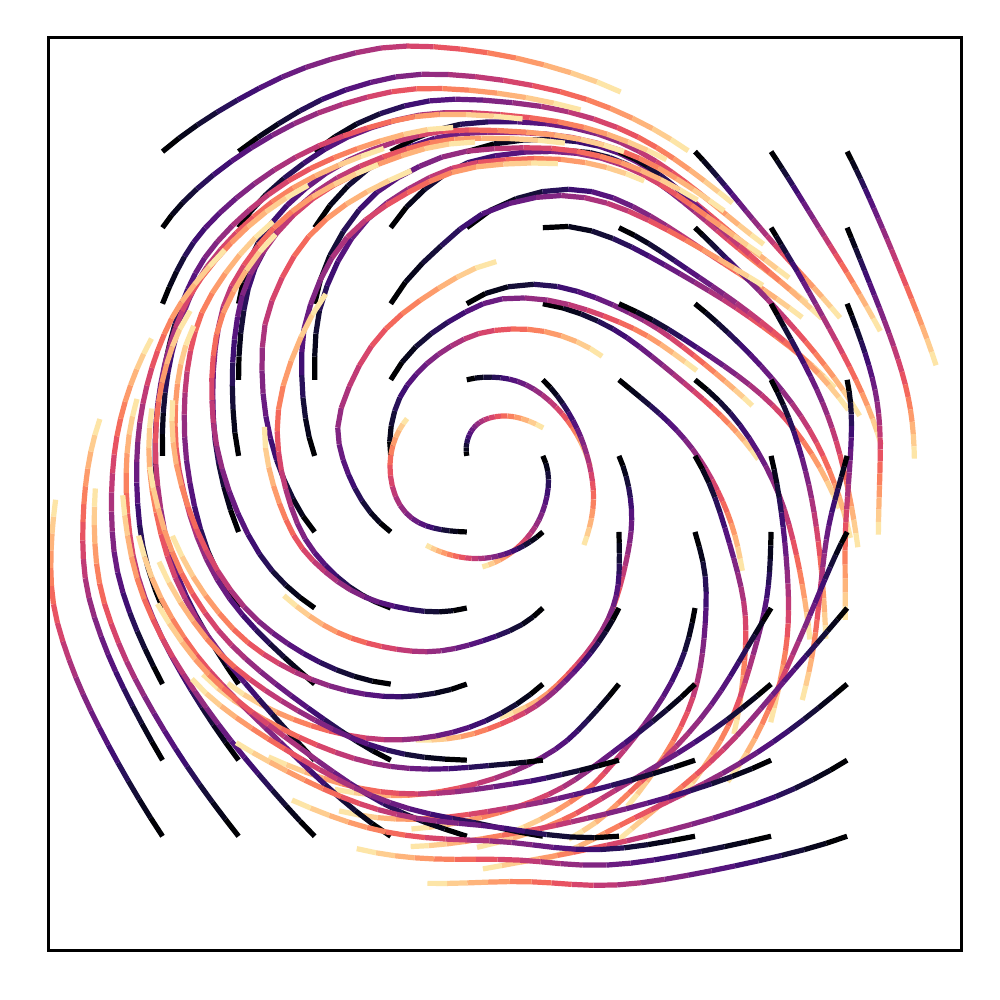}
        \includegraphics[width=\picsize]{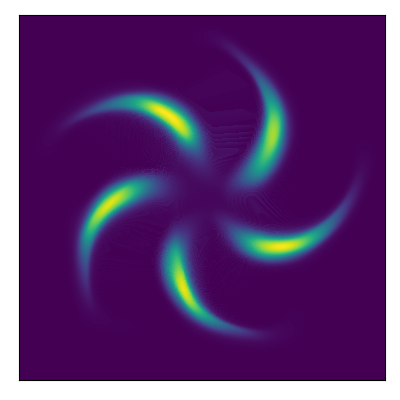}
        \includegraphics[width=\picsize]{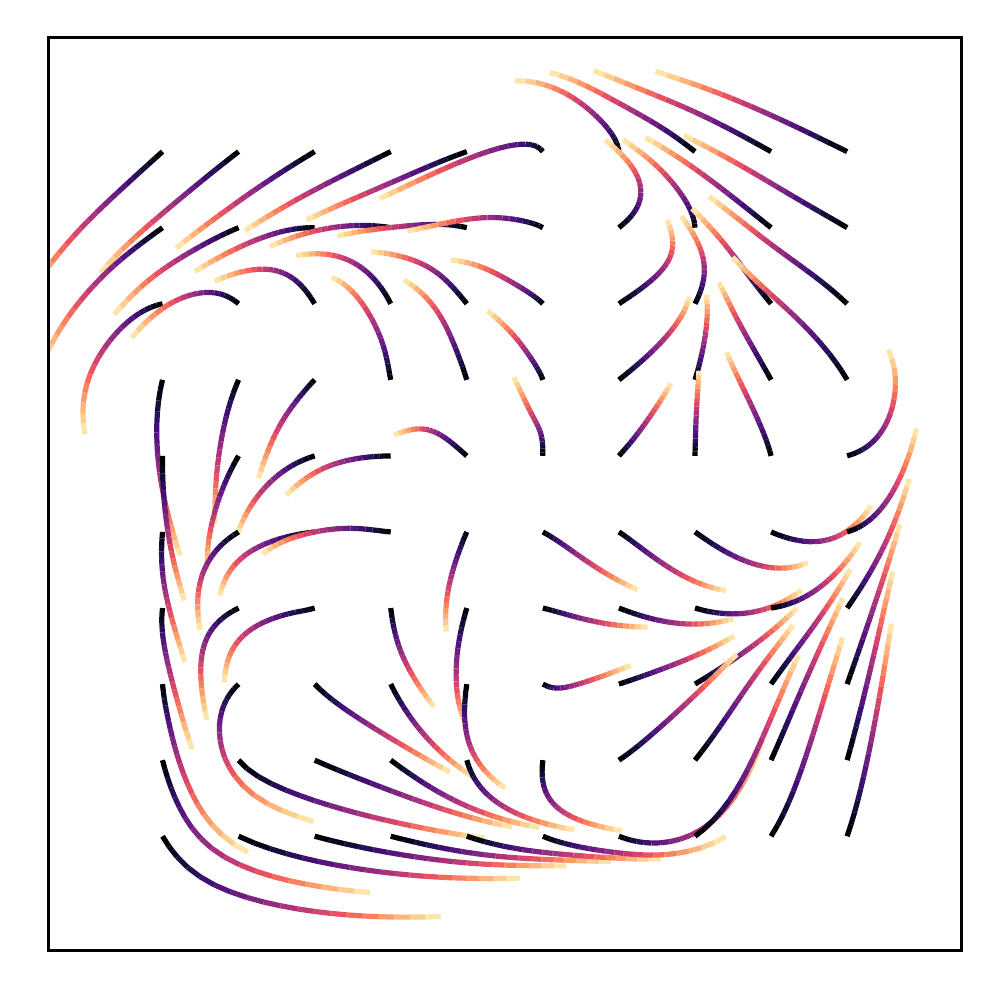}
        \includegraphics[width=\picsize]{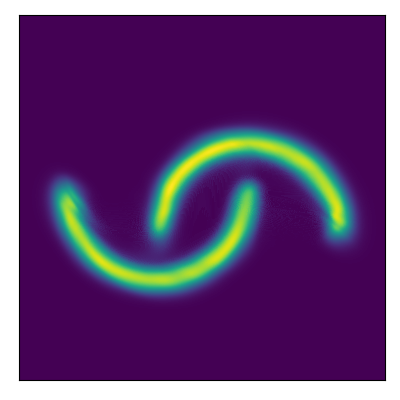}
        \includegraphics[width=\picsize]{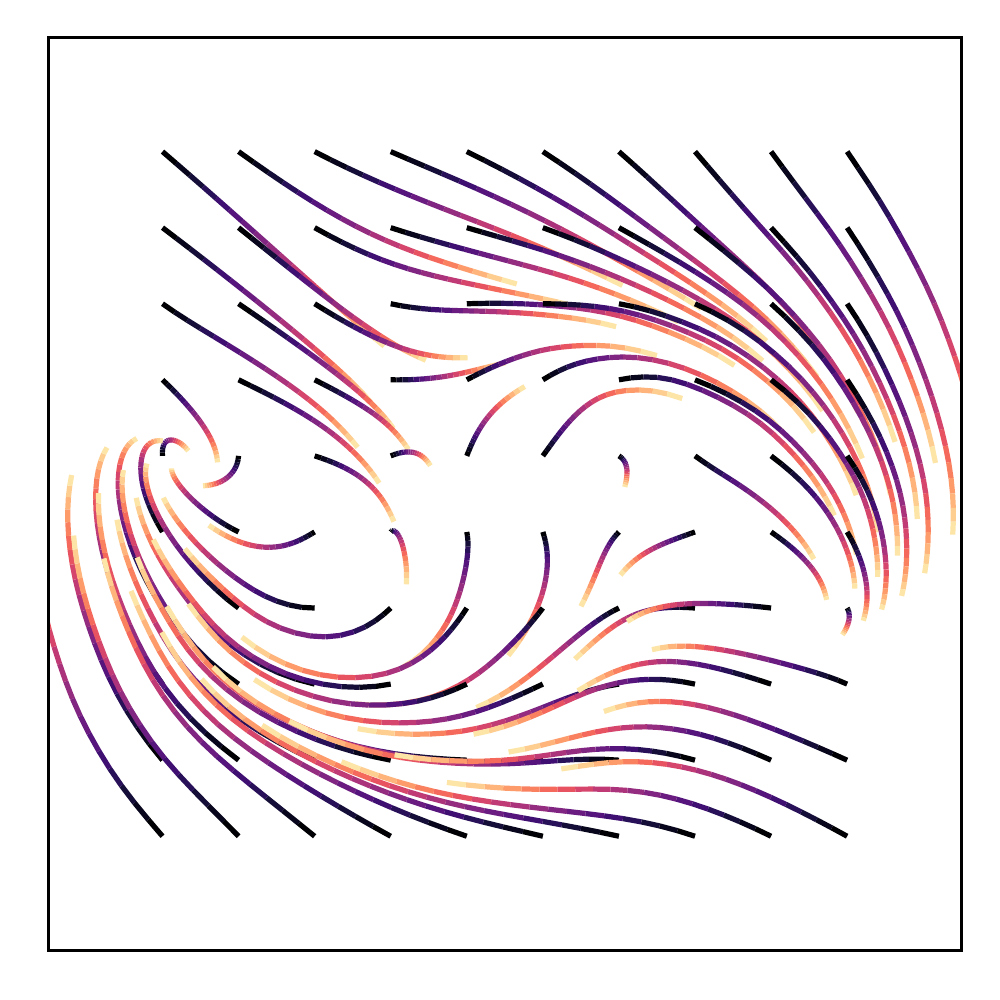}
        \caption{FFJORD}
        \label{subfig:2d-ffjord}
    \end{subfigure}
    \begin{subfigure}{0.33\linewidth}
        \centering
        \includegraphics[width=\picsize]{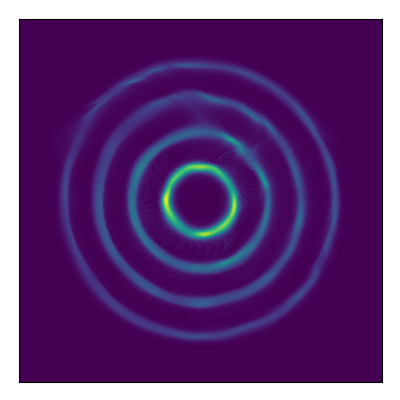}
        \includegraphics[width=\picsize]{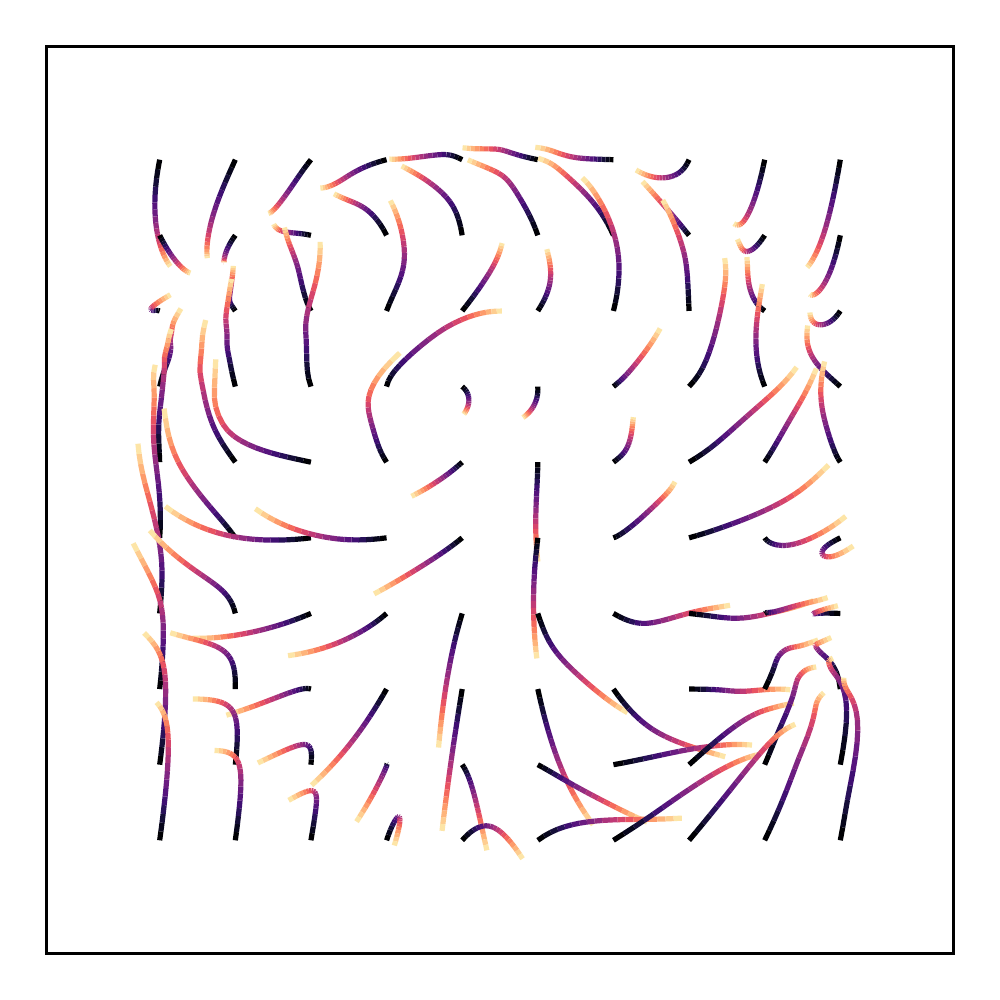}
        \includegraphics[width=\picsize]{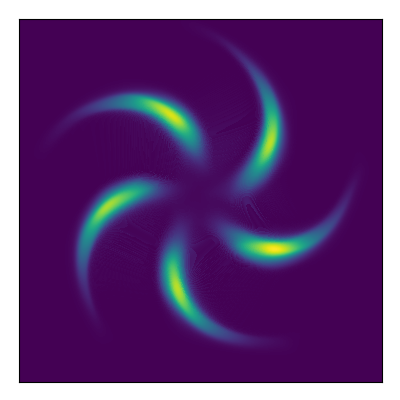}
        \includegraphics[width=\picsize]{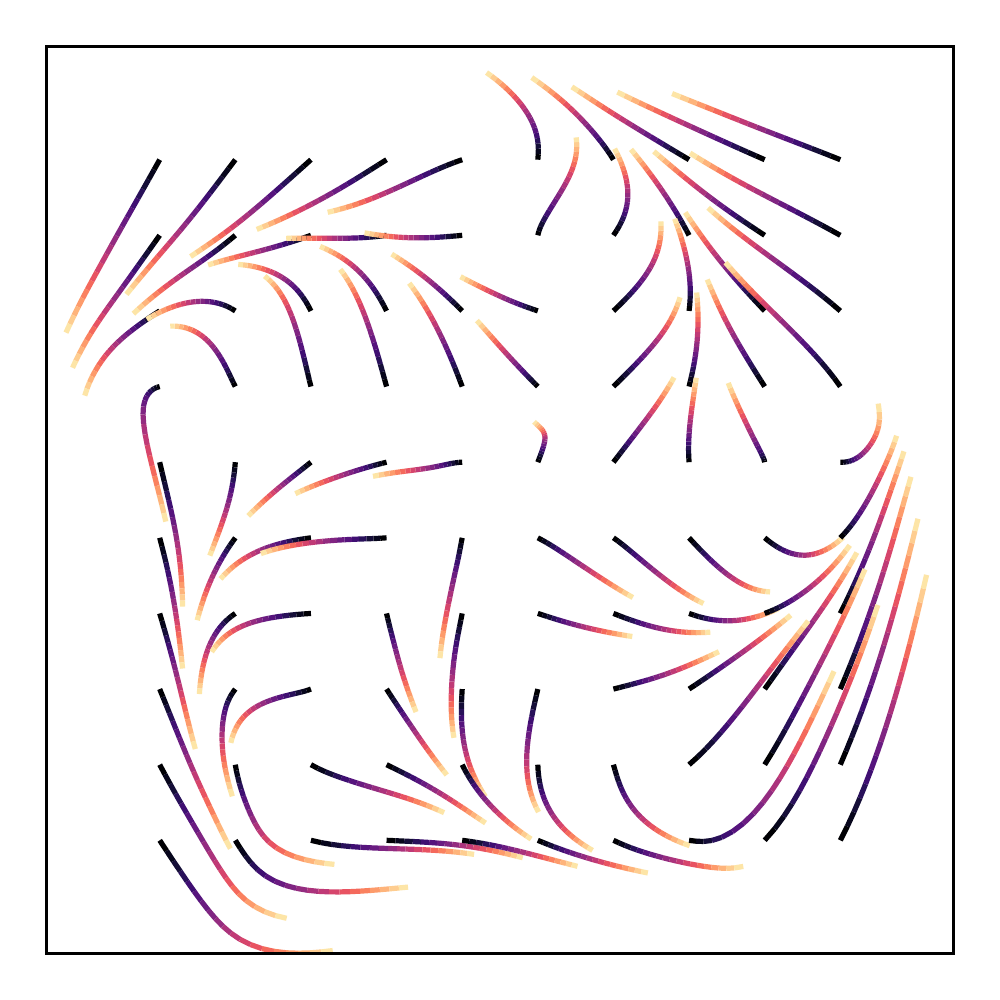}
        \includegraphics[width=\picsize]{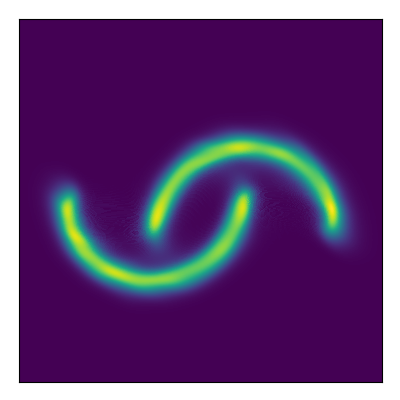}
        \includegraphics[width=\picsize]{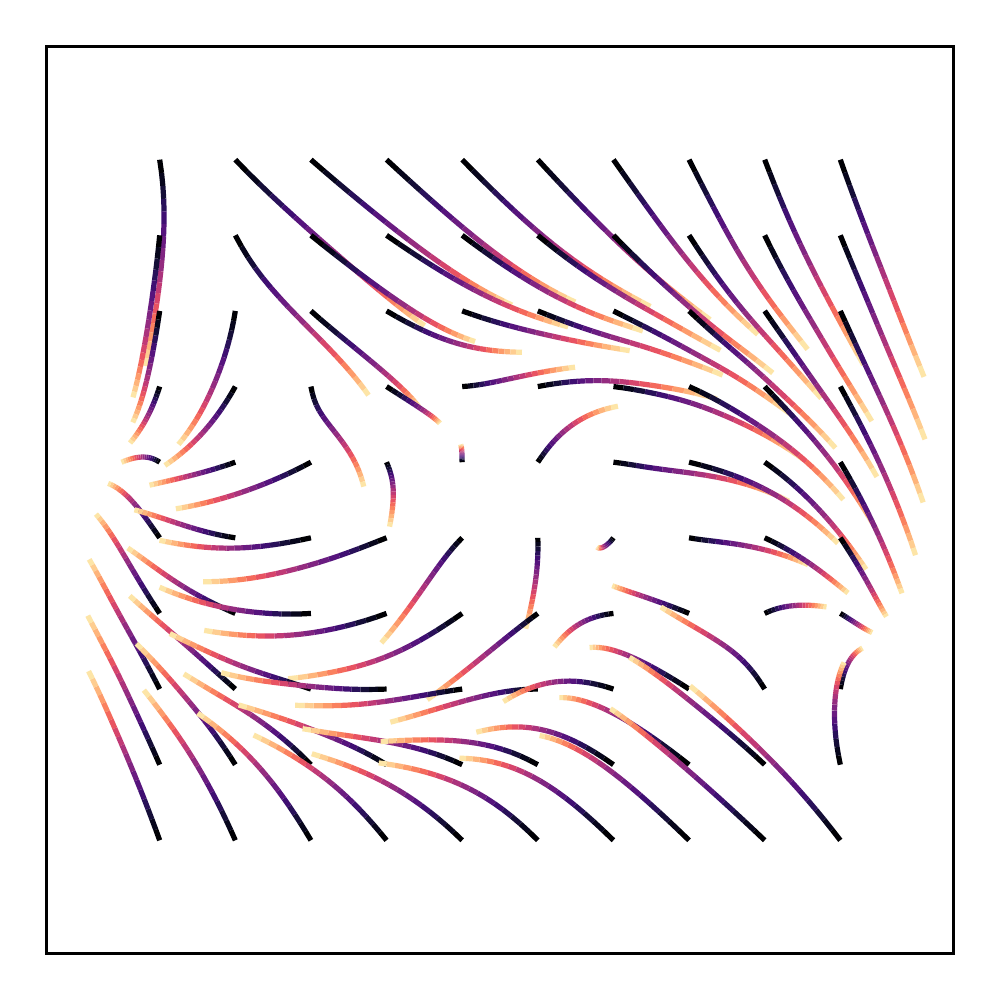}
        \caption{Ours}
        \label{subfig:2d-ours}
    \end{subfigure}
    \caption{\subref{subfig:2d-data}: Distribution of three 2D data sets. \subref{subfig:2d-ffjord} and \subref{subfig:2d-ours}: The reconstructed distributions and the trajectories of $\vb{y}(t)$ by FFJORD and our model. To generate the trajectories, we select some grid points to be the initial values $\vb{y}(t_0)$ and solve the ODE in Equation~\eqref{eq:dynamics_y} to obtain $\vb{y}(t)$. Each curve in the figures represents a single trajectory of $\vb{y}(t)$, and the color represents the value of time $t$.}
    \label{fig:2d-data}
\end{figure*}

In this experiment, we test with three simple 2D toy data.
The distributions of the three 2D data used for the experiment are shown in Figure \ref{subfig:2d-data}. 
The approximated distributions using FFJORD and our model are in the upper halves of Figures \ref{subfig:2d-ffjord} and \ref{subfig:2d-ours}, respectively. 
Both FFJORD and our model successfully recover the data distributions in decent quality, which implies that our new regularization loss doesn't harm approximation capability, and the model is able to find a good solution.

We visualize the trajectories of $\vb{y}(t)$ in the lower halves of Figures \ref{subfig:2d-ffjord} and \ref{subfig:2d-ours}. 
We want to know whether our model actually performs how we expected as in Figure \ref{subfig:withTPR}, which is that the trajectories of $\vb{y}(t)$ are approximately straight lines.
It can be seen that the trajectories of FFJORD are very winding, while our trajectories are less winding and most are straight.
Since we are using polynomial regularization of degree 1, the result is what we expected.
Because of this, our ODEs are simpler to solve and the required step sizes for solving ODEs can be larger, so the number of steps can be smaller in training.
In reality, the average NFEs of FFJORD on the three data are $90.42$, $60.48$ and $86.7$ respectly, while the NFE of our model are just $44.17$, $47.39$ and $33.66$. That is, our TPR leads to respectively $51.15\%$, $21.64\%$ and $60.90\%$ drop of computational cost.

\subsubsection{Results on real data}
\begin{table*}
    \centering
    \caption{The average NFE and testing negative log-likelihood on tabular data (in nats) and image data (in bits/dim). FFJORD represents the results reported in the original paper, while FFJORD* is the results run by us. The value of NFE is not reported in the original paper of FFJORD.}
    \begin{tabular}{lccccccc}
        \toprule
            & POWER & GAS & HEPMASS & MINIBOONE & BSDS300 & MNIST & CIFAR10\\
        \midrule
            & \multicolumn{7}{c}{Average NFE}\\
        \midrule
            FFJORD*  & 885.19 & 488.00 & 628.41 & 107.29 & 284.51 & 399.01 & 530.41 \\
            Ours & \textbf{253.88} & \textbf{178.08} & \textbf{260.14} & \textbf{32.76} & \textbf{87.05} & \textbf{221.09} & \textbf{306.12}\\
        \midrule
            & \multicolumn{7}{c}{Testing Loss} \\
        \midrule
            Real NVP & -0.17 & -8.33 & 18.71 & 13.55 & -153.28 & 1.06 & 3.49\\
            Glow     & -0.17 & -8.15 & 18.92 & 11.35 & -155.07 & 1.05 & \textbf{3.35}\\
            FFJORD & -0.46 & -8.59 & \textbf{14.92} & \textbf{10.43} & -157.40 & 0.99 & 3.40\\
            FFJORD* & -0.46 & \textbf{-11.56} & 15.79 & 11.37 & -157.46 & 0.96 & \textbf{3.35} \\
            Ours & \textbf{-0.50} & -11.36 & 15.85 & 11.34 & \textbf{-157.94} & \textbf{0.95} & 3.36\\
        \midrule
            MADE   & 3.03 & -3.56 & 20.98 & 15.59 & -148.85 & 2.04 & 5.67 \\
            MAF    & -0.24 & -10.08 & 17.70 & 11.75 & -155.69 & 1.89 & 4.31 \\
            TAN    & -0.48 & -11.19 & 15.12 & 11.01 & -157.03 & 1.19 & 3.98 \\
            MAF-DDSF & -0.62 & -11.96 & 15.09 & 8.86 & -157.73 & - & - \\
        \bottomrule
    \end{tabular}
    \label{tab:density_results}
\end{table*}

The real data we used here are tabular data, which are five data sets processed in \citep{papamakarios2017masked}, and the image data including the MNIST and CIFAR10 data sets.
Besides FFJORD, we compare our model with two  discrete normalizing flows:  Real NVP \citep{dinh2016density} and Glow \citep{kingma2018glow}, and four auto-regressive flows: MADE \citep{kingma2016improved}, MAF~\citep{papamakarios2017masked}, TAN~\citep{oliva2018transformation} , and MAF-DDSF~\cite{huang2018neural}.
But the main purpose of our work is improving continuous normalizing flow, so we mainly compare our result with FFJORD.

For tabular data, we use the same neural network architectures as those in FFJORD, and for image data, the same multiscale~\citep{dinh2016density} architectures as in FFJORD are used.
For image data, due to our hardware limitation, we used a smaller batch size (200) and smaller number of total training epoch (200). Despite these adjustments, we find the training quality of FFJORD is not affected. For tabular data the setup is the same as those in FFJORD.

The results of average NFE and negative log-likelihood on testing data are shown in Table~\ref{tab:density_results}. 
For average NFE, our model significantly outperforms FFJORD. On all data sets, our model uses fewer NFE in training. The reduction of NFE ranges from 42.3\% to 71.3\%. The largest reduction, 71.3\%, occurs on POWER data set.
Note that the actually training time is approximately proportional to NFE. Therefore, our model takes significantly less time in training.

In terms of testing loss, our model is comparable to the original FFJORD. 
It can be seen that our testing loss is very similar to that of FFJORD*, which is the FFJORD model run by us. 
Our model even produces lower testing loss on four out of seven data sets than FFJORD*.
This result matches our expectation that the additional TPR loss does not affect the performance much, as discussed in Section~\ref{sec:method}.
After all, our purpose is improving training efficiency but not testing loss. 

\begin{figure*}
    \centering
    \begin{subfigure}{0.65\linewidth}
    \centering
    \includegraphics[width=0.49\linewidth]{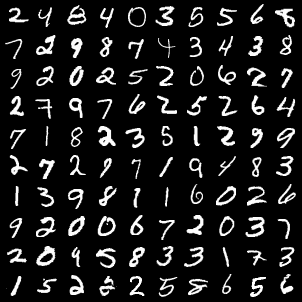}
    \includegraphics[width=0.49\linewidth]{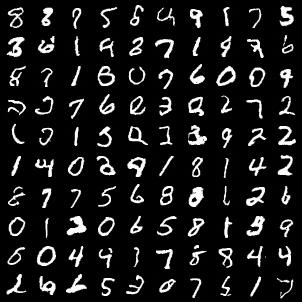}
    \caption{Left: Ground truth images of MNIST data set. Right: Generated images by our model trained on MNIST data set.}
    \end{subfigure}
    \centering
    \begin{subfigure}{0.65\linewidth}
    \centering
    \includegraphics[width=0.49\linewidth]{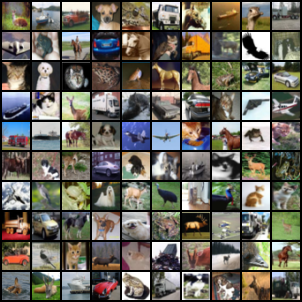}
    \includegraphics[width=.49\linewidth]{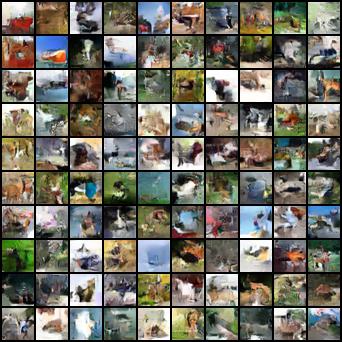}
    \caption{Left: Ground truth images of CIFAR10 data set. Right: Generated images by our model trained on CIFAR10 data set..}
    \end{subfigure}
    \caption{The true images and the generated images by our model on two image data sets, MNIST and CIFAR10. The images are generated by running the CNF model in reverse with Gaussian noise as input.}
    \label{fig:generated_images}
\end{figure*}

For image data sets, MNIST and CIFAR10, our model also works great. The two data sets have the highest dimension of data. Therefore, they are the most difficult to train among all these data sets. The fact that our model yields good results on the two data sets indicates that our model can work on difficult data with high dimensions. For the two image data sets, the generated images sampled from our trained model are shown in Figure~\ref{fig:generated_images}.

\begin{table*}
    \centering
    \caption{The average NFE and testing negative ELBO, with mean and stdev, for VAE models on four data sets. FFJORD represents the results reported in the original paper, while FFJORD* is the results run by us. The value of NFE is not reported in the original paper of FFJORD}
    \begin{tabular}{lcccc}
        \toprule
            & MNIST & Omniglot &  Frey Faces & Caltech Silhouettes  \\
        \midrule
             & \multicolumn{4}{c}{Average NFE} \\
        \midrule
             FFJORD* & 58.72 $\pm$ 1.32 & 108.73 $\pm$ 5.15 & 72.19 $\pm$ 6.96 & 39.83 $\pm$ 2.56\\
             Ours & \textbf{40.45 $\pm$ 0.37} & \textbf{87.73 $\pm$ 2.59} & \textbf{52.71 $\pm$ 0.87} & \textbf{27.06 $\pm$ 0.51} \\
        \midrule
             & \multicolumn{4}{c}{Testing Loss} \\
        \midrule
             No Flow & 86.55 $\pm$ .06 & 104.28 $\pm$ .39 & 4.53 $\pm$ .02 & 110.80 $\pm$ 0.46 \\
             Planar  & 86.06 $\pm$ .31 & 102.65 $\pm$ .42 & 4.40 $\pm$ .06 & 109.66 $\pm$ 0.42 \\
             IAF     & 84.20 $\pm$ .17 & 102.41 $\pm$ .04 & 4.47 $\pm$ .05 & 111.58 $\pm$ 0.38 \\
             Sylvester & 83.32 $\pm$ .06 & 99.00 $\pm$ .04 & 4.45 $\pm$ .04 & 104.62 $\pm$ 0.29\\
             FFJORD & 82.82 $\pm$ .01 & 98.33 $\pm$ .09 & \textbf{4.39 $\pm$ .01} & 104.03 $\pm$ 0.43\\
             FFJORD* & \textbf{81.90 $\pm$ .08} & \textbf{97.40 $\pm$ .21} & 4.49 $\pm$ .04 & 102.84 $\pm$ 1.31\\
             Ours & 81.94 $\pm$ .06 & 97.52 $\pm$ .02 & 4.51 $\pm$ .08 & \textbf{102.12 $\pm$ 0.56} \\
         \bottomrule
    \end{tabular}
    \label{tab:vae_results}
\end{table*}
\subsection{Variational autoencoder (VAE)}

We perform VAE experiments on four data sets obtained from \citep{berg2018sylvester}.
The loss of this task is the negative evidence lower bound (ELBO).
\begin{equation}
    L_0(\vb{x})=\mathrm{D_{KL}}(q_\phi(\vb{z}|\vb{x})\|p(\vb{z}))
    -\expectation_{q_\phi(\vb{z}|\vb{x})}[ \log p_\theta(\vb{x}|\vb{z})]. 
\end{equation}
In this experiment, we further compare the state-of-the-art discrete models, including Planar Flow \citep{rezende2015variational}, Inverse Autoregressive Flow (IAF) \citep{kingma2016improved} and Sylvester normalizing flow \citep{berg2018sylvester}.

The hyper-parameters of neural network architecture and training of our model are the same as those in FFJORD.
In short, the neural network is composed of the low-rank parameter encoding layers, to encode the input data into CNF.
The learning rate is set to be $5\times 10^{-4}$ and is divided by 10 when the validation error does not decrease for 35 epochs. Each model is run three times in experiment.

The result of average training NFE and testing negative ELBO is shown in Table~\ref{tab:vae_results}. For average NFE, our model still significantly outperforms FFJORD. 
Our model has reduced NFE ranging from 19.3\% to 32.1\%. The largest reduction is on the Caltech Silhouettes data set.
For the negative ELBO, our model performs very similarly to FFJORD. This exactly meets our expectation that simply adding TPR reduces NFE but keeps testing loss the same.

\section{Conclusion}\label{conclusion}
We have improved the computational efficiency of continuous normalizing flow models. High computational cost was the largest limitation of CNF.
We proposed to  add a loss function based on polynomial regression to regularize the trajectory shape of the ODE solution. Our method is proposed to reduce the truncation errors of solving ODE and can result in fewer NFE. Furthermore, we proved that with this regularization, there are always optimal solutions of the vector field for the original loss function, and we argued that our new regularization doesn't harm testing loss. 
Empirically, our model reduces a great amount of computation cost, while reaching a comparable testing to FFJORD.

\section*{Acknowledgments}
This study was supported in part by the Ministry of Science and Technology (MOST) of Taiwan, R.O.C., under Contracts 107-2221-E-001-009-MY3 and 106-3114-E-002-008.

\bibliography{reference}
\clearpage

\appendix

\section{Proofs}
\label{appendix:proofs}
\subsection{Proof of Theorem \ref{thm:existence}}
To prove Theorem \ref{thm:existence}, we construct a vector field $\vb{v}(\vb{y}, t)$ to satisfy the two conditions. 
The first one is $\dv{\vb{v}}{t}=0$ for all $\vb{y}$ and $t$. 
And the second one is the initial and final values of $\log p(t)$.

Let $\vb{s}(\vb{x}, t)$ be the solution of the differential equation \eqref{eq:dynamics_y} in Section \ref{sec:background}, with $\vb{x}$ being its initial value, i.e.\ $\vb{s}(\vb{x}, t_0) = \vb{x}$. 
Define $\vb{z}(\vb{x}) = \vb{s}(\vb{x}, t_1) $ be the end value of the solution.
Because $\dv{\vb{v}}{t} = 0$, $\vb{s}$ is moving in space at a constant velocity from $\vb{x}$ to $\vb{z}$, so we can deduce that
\begin{equation}
    \vb{s}(\vb{x}, t) = \vb{x} + \frac{t-t_0}{t_1-t_0}\left(\vb{z}(\vb{x}) - \vb{x}\right).
    \label{eq:thm1_s}
\end{equation}
Once the trajectories $\vb{s}(\vb{x},t)$ are all defined, the vector field $\vb{v}(\vb{y}, t)$ is also determined.
Consequently, the goal to find the vector field $\vb{v}(\vb{y}, t)$ becomes finding the mapping $\vb{z}(\vb{x})$.

Note that $\vb{v}(\vb{y}, t)$ should be well-defined at any point.
For example, the function $\vb{z}(\vb{x}) = -\vb{x}$ will lead to an ill-defined $\vb{v}$, because at $t=\frac{t_0+t_1}{2}$, $\vb{s}$ is at the origin no matter what its initial value $\vb{x}$ is, and thus $\vb{v}(0, \frac{t_0+t_1}{2})$ can not be defined.

To make $\vb{v}(\vb{y}, t)$ well-defined, we must find the requirement.
Because $\vb{v}$ is constant on the trajectory $\vb{s}(\vb{x}, t)$, we have the equation
$\vb{v}(\vb{s}(\vb{x},t), t) = \vb{v}(\vb{x}, t_0)$. 
It can be seen that $\vb{v}(\vb{s}(\vb{x},t), t)$ is well-defined if there is only one unique $\vb{x}$ that traverses to $\vb{s}(\vb{x},t)$ at time $t$.
In other words, the inverse function $\vb{x}=\vb{s}^{-1}(\vb{y}, t)$ needs to be well-defined.
From the inverse function theorem, $\vb{s}(\vb{x},t)$ is invertible if the determinant of Jacobian is not zero, i.e.\ $\det \pdv{\vb{s}(\vb{x}, t)}{\vb{x}} \ne 0$ for all $\vb{x}$ and $t$. 
Substituting $\vb{s}$ with equation \eqref{eq:thm1_s}, the requirement of determinant of Jacobian becomes
\begin{equation}
    \det((1-\xi) I + \xi \pdv{\vb{z}}{\vb{x}}) \ne 0,\quad \text{where } 0 \le \xi\equiv\frac{t-t_0}{t_1-t_0} \le 1.
    \label{eq:thm1_condition_v}
\end{equation}

The second condition to prove the theorem is to satisfy the initial and final conditions of $\log p$. 
Given the transformation $\vb{z} = \vb{z}(\vb{x})$,
we have the equation for probabilities, $p(\vb{x}) = p(\vb{z})\abs{\det\pdv{\vb{z}}{\vb{x}}}$. Since the initial value condition is $p(\vb{x}) = p_0(\vb{x})$ and the final value condition is $p(\vb{z}) = p_1(\vb{z})$, we must have the equation below for $\vb{z}(\vb{x})$, 
\begin{equation}
    p_0(\vb{x}) = p_1(\vb{z})\abs{\det\pdv{\vb{z}}{\vb{x}}}.
    \label{eq:thm1_condition_logp}
\end{equation}

To prove Theorem \ref{thm:existence}, we have to find at least one $\vb{z}(\vb{x})$ such that both conditions \eqref{eq:thm1_condition_v} and \eqref{eq:thm1_condition_logp} are satisfied. 
Our trick is to assume that the function $\vb{z}(\vb{x})$ satisfies the equations
\begin{equation}
  \begin{aligned}
    z_1 &= z_1(x_1), \\
    z_2 &= z_2(x_1, x_2), \\
    &\quad\vdots \\
    z_d &= z_d(x_1, x_2,\dots, x_d), \\
  \end{aligned}
\end{equation} 
where $x_i$ and $z_i$ are the $i$-th dimensions of $\vb{x}$ and $\vb{z}$.
With these equations, the Jacobian matrix $\pdv{\vb{z}}{\vb{x}}$ is simply a lower triangular matrix, and then its determinant is simply $\prod_i \pdv{z_i}{x_i}$.

\paragraph{Satisfaction of equation \eqref{eq:thm1_condition_logp}}
Since $p_0$ and $p_1$ can be written as the product of conditional probabilities,
\begin{equation}
  \begin{aligned}
    p_0(\vb{x})&=p_0(x_1)p_0(x_2|x_1)\dots p_0(x_d|x_1,\dots,x_{d-1}), \\
    p_1(\vb{z})&=p_1(z_1)p_1(z_2|z_1)\dots p_1(z_d|z_1,\dots,z_{d-1}),
  \end{aligned}
\end{equation} 
we can construct $\vb{z}(\vb{x})$ by assuming that it follows the differential equations 
\begin{equation}
  \begin{aligned}
    \pdv{z_1}{x_1} &= \frac{p_0(x_1)}{p_1(z_1)}, \\
    \pdv{z_2}{x_2} &= \frac{p_0(x_2|x_1)}{p_1(z_2|z_1)}, \\
    &\vdots \\
    \pdv{z_d}{x_d} &= \frac{p_0(x_d|x_{1:d-1})}{p_1(z_d|z_{1:d-1})}, \\
  \end{aligned}
\end{equation} 
where $x_{1:d-1}$ denotes the vector $(x_1, x_2, \dots ,x_{d-1})$.
After these equation are solved, $\vb{z}(\vb{x})$ can be determined.
And due to these equations, the determinant of Jacobian becomes
\begin{equation}
  \det\pdv{\vb{z}}{\vb{x}}
  =\prod_{i=1}^d \frac{p_0(x_i|x_{1:i-1})}{p_1(z_i|x_{1:i-1})}
  = \frac{p_0(\vb{x})}{p_1(\vb{z})}.
\end{equation}
Thus, condition \eqref{eq:thm1_condition_logp} is satisfied.

\paragraph{Satisfaction of inequality \eqref{eq:thm1_condition_v}}
Since $\pdv{\vb{z}}{\vb{x}}$ is a triangular matrix,  the left hand side of equation \eqref{eq:thm1_condition_v} becomes
\begin{equation}
  \begin{aligned}
    \det((1-\xi) I + \xi \pdv{\vb{z}}{\vb{x}}) 
    &= \prod_{i=1}^d\left((1-\xi) + \xi \pdv{z_i}{x_i}\right)
    \\
    &= \prod_{i=1}^d\left((1-\xi) + \xi \frac{p_0(x_i|x_{1:i-1})}{p_1(z_i|z_{1:i-1})}\right)
    \\
    &>0.
  \end{aligned}
\end{equation}
The determinant is always greater than zero, so condition \eqref{eq:thm1_condition_v} is satisfied.

Because we can construct at least one solution $\vb{v}(\vb{y}, t)$ to satisfy all the conditions, Theorem \ref{thm:existence} is proved.

\subsection{Proof of Theorem \ref{thm:infinity}}
If we can find any infinitesimal function $\delta\vb{z}(\vb{x})$ such that $\vb{z}(\vb{x}) + \delta\vb{z}(\vb{x})$ still satisfies the
conditions \eqref{eq:thm1_condition_v} and \eqref{eq:thm1_condition_logp}, 
then we can arbitrarily construct another solution of $\vb{v}$ based on the solutions found. 
And thus there are infinitely many solutions of $\vb{v}$. 

Assume that
\begin{equation}\label{eq:zprime}
  \vb{z}'(\vb{x}, \epsilon)=\vb{z}(\vb{x})+\epsilon \vb{v}'(\vb{z}(\vb{x})),
\end{equation} 
where $\vb{z}(\vb{x})$ is a solution that satisfies \eqref{eq:thm1_condition_v} and \eqref{eq:thm1_condition_logp}, and $\epsilon$ is an infinitesimal variable.
To make $\vb{z}'(\vb{x},\epsilon)$ still a solution, we need to find $\vb{v}'$ such that $\vb{z}'$ also satisfies \eqref{eq:thm1_condition_v} and \eqref{eq:thm1_condition_logp}. 

First, the left hand side of condition \eqref{eq:thm1_condition_v} now becomes
\begin{equation}
  \det((1-\xi)I+\xi\pdv{\vb{z}}{\vb{x}}\left(I+\epsilon\pdv{\vb{v}'}{\vb{z}}\right)).
\end{equation}
To satisfy condition \eqref{eq:thm1_condition_v}, we assume that every element of the Jacobian matrix $\pdv{\vb{v}'}{\vb{z}}$ is bounded. 
If they are bounded, we can always find an $\epsilon$ sufficiently small such that the overall determinant still stays unequal to zero.

For the condition \eqref{eq:thm1_condition_logp}, we don't directly use equation \eqref{eq:thm1_condition_logp} to prove.
Instead, we assume that $\log p(\vb{z},\epsilon)$ is the log probability with the points in space following the trajectory in equation \eqref{eq:zprime}. 
So $\log p(\vb{z}')=\log p(\vb{z}'(\vb{x},\epsilon), \epsilon)$. Because we have $\log p(\vb{z}, 0) = \log p_1(\vb{z})$ and we need that $\log p(\vb{z}')=\log p_1(\vb{z}')$, we can obtain the equation below, 
\begin{equation}
    \eval{\pdv{}{\epsilon}\log p(\vb{z}, \epsilon)}_{\epsilon=0} = 0.
\end{equation}
On the other hand, from the dynamics of $\log p$ in equation \eqref{eq:dynamics_logp}, we also know that
\begin{equation}
  \eval{\dv{}{\epsilon}\log p(\vb{z}')}_{\epsilon=0}=  -\nabla_{\vb{z}}\cdot\vb{v}',
\end{equation} 
where $\nabla_{\vb{z}}\cdot\vb{v}'\equiv\Tr(\pdv{\vb{v}'}{\vb{z}})$ is called the divergence of $\vb{v}'$.
Below, we will omit the subscript $\vb{z}$ in $\nabla_{\vb{z}}$ and just use $\nabla$.
From the chain rule, 
$\eval{\dv{\log p}{\epsilon}}_{\epsilon=0} = \eval{\pdv{\log p}{\epsilon}}_{\epsilon=0} + \vb{v}'\vdot\grad\log p_1$, 
we can derive the following equation 
\begin{equation}
  \vb{v}'\vdot\grad\log p_1 + \divergence\vb{v}' = 0.
  \label{eq:thm2_vprime}
\end{equation}

To find a $\vb{v}'(\vb{z})$ for  equation \eqref{eq:thm2_vprime}, let us first assume that $\vb{v}'$ has the form 
\begin{equation}
  \vb{v}'(\vb{z})=e^{g(\vb{z})}\vb{u}(\vb{z}).
\end{equation} 
Substituting it into equation \eqref{eq:thm2_vprime}, we get
\begin{equation}
  \vb{u}\vdot\grad\log p_1 + \divergence\vb{u} + \vb{u}\vdot\grad g = 0.
\end{equation} 
Because $g(\vb{z})$ can be any function, we can set $g = -\log p_1$, and obtain the equation 
\begin{equation}
  \divergence \vb{u} =0.
\end{equation}
And thus 
\begin{equation}
    \vb{v}' = \frac{\vb{u}}{p_1}.
\end{equation}

It may seem that we have completed the proof because there are infinite solutions for the equation $\divergence\vb{u}=0$.
However, most of the solutions of $\vb{u}$ do not guarantee that the Jacobian matrix $\pdv{\vb{v}'}{\vb{z}}$ is bounded.
Since $p_1$ asymptotically tends to $0$ at infinity, if $\vb{u}$ does not decay faster than $p_1$, $\vb{v}'$ and its Jacobian will explode when approaching infinity.
This is why this proof doesn't work for $d=1$, because  $\divergence\vb{u}=0$ implies that $\vb{u}$ is a constant and then should be $0$.

If we can find a $\vb{u}(\vb{z})$ that is nonzero only in a bounded area, the problem can be solved.
We can see that it is possible by imagining an incompressible fluid confined in a box.
Because of incompressibility, the velocity field of the fluid has zero divergence everywhere inside the box.
In addition, the velocity field outside the box is always zero.
Therefore, the velocity field is nonzero only in the box and has zero divergence everywhere.

Mathematically, we can set $\vb{u}(\vb{z})$ as
\begin{equation}
  u_i = -c_i \frac{L_i}{\pi}\left(1+\cos(\frac{z_i \pi}{L_i})\right)\prod_{j\ne i}\sin(\frac{z_j \pi}{L_j})
\end{equation} 
inside the box $-L_i\le z_i \le L_i$, for $i= 1,2,\dots, d$, and $\vb{u}=0$ outside the box.
It can be seen that $\vb{u}$ is continuous everywhere,
and it has the divergence 
\begin{equation}
  \divergence\vb{u} = \left(\sum_i c_i\right)\prod_i \sin(\frac{z_i \pi}{L_i}) 
\end{equation} 
inside the box.
For $d>1$, we can always find at least one set of  $c_i$ so that $\sum_i c_i = 0$ and thus $\divergence\vb{u} = 0$ everywhere. 
Consequently, because of the existence of $\vb{u}$ and $\vb{v}'$, the proof is completed.

\end{document}